\documentclass[letterpaper, 10 pt, conference]{ieeeconf}  %
\IEEEoverridecommandlockouts                              

\usepackage{amsmath} 
\usepackage{amssymb}  
\usepackage{graphicx}
\usepackage{soul}
\usepackage{xcolor}
\usepackage[acronym]{glossaries}
\usepackage[ruled,vlined,noend]{algorithm2e}
\usepackage{algpseudocode}%
\usepackage{hyperref}
\usepackage{multicol}
\usepackage{cite}
\usepackage{siunitx}
\usepackage{float}
\usepackage{stfloats}
\usepackage{bm}
\usepackage{url}
\title{\LARGE \bf
An Adaptive Inspection Planning Approach Towards Routine Monitoring in Uncertain Environments
}

\author{Vignesh Kottayam Viswanathan, Yifan Bai, Scott Fredriksson, Sumeet Satpute,\\ Christoforos Kanellakis and  George Nikolakopoulos
\thanks{The authors are with Robotics and AI, Luleå University of Technology 97187, Luleå,Sweden
        {\tt\small \{vigkot, yifan.bai, scofre, sumsat, chrkan and geonik\}@ltu.se}}%
\thanks{The project is funded by the European Union's Horizon Europe Research and Innovation Programme, under the Grant Agreement No. 101138451 PERSEPHONE}
}

\begin{document}

\maketitle
\thispagestyle{empty}
\pagestyle{empty}


\begin{abstract}
In this work, we present a hierarchical framework designed to support robotic inspection under environment uncertainty. By leveraging a known environment model, existing methods plan and safely track inspection routes to visit points of interest. However, discrepancies between the model and actual site conditions, caused by either natural or human activities, can alter the surface morphology or introduce path obstructions. To address this challenge, the proposed framework divides the inspection task into: (a) generating the initial global view-plan for region of interests based on a historical map and (b) local view replanning to adapt to the current morphology of the inspection scene. The proposed hierarchy preserves global coverage objectives while enabling reactive adaptation to the local surface morphology. This enables the local autonomy to remain robust against environment uncertainty and complete the inspection tasks. We validate the approach through deployments in real-world subterranean mines using quadrupedal robot. A supplementary media highlighting the proposed method can be found here \url{https://youtu.be/6TxK8S_83Lw}.
\end{abstract}
\section{Introduction}
\begin{figure}[htbp]
    \centering
    \includegraphics[width=\linewidth]{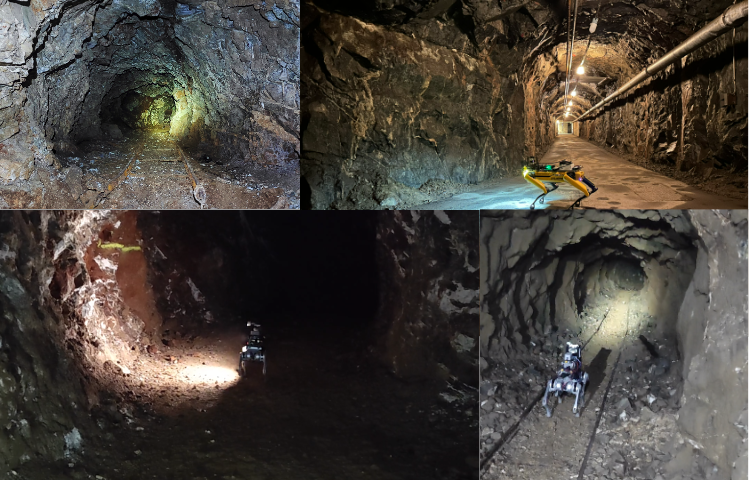}
    \caption{Examples of robotic inspection application in subterranean environments. }
    \label{fig:concept_examples}
\end{figure}

Robotic inspection plays a critical role in modern mining operations, with robots increasingly deployed to handle unsafe and repetitive tasks: from detecting changes in subterranean environments~\cite{stathoulopoulos2023irregular, attard2018vision} to routine monitoring~\cite{staniaszek2025autoinspect,nordstrom2024safety}. This work contributes to enabling autonomous visual inspection in uncertain subterranean environments, focusing specifically on surface inspection and the discrepancies that arise between a priori models and current environmental conditions. These discrepancies may result from obstructions along pre-recorded paths (e.g., rockfalls) or from routine mining activities that alter surface morphology (e.g., ore extraction). Since such uncertainty is often unpredictable and cannot be fully modelled in advance, it can degrade inspection performance and compromise the quality of visual data collected during missions. Figure~\ref{fig:concept_examples} presents a few demonstrative use-case of robotic inspection application in subterranean mines. 

Therefore, addressing visual inspection tasks in a potentially outdated environment model poses two main challenges. First, discrepancies between current conditions and historical environment data can compromise the generated inspection route. These mismatches may introduce obstacles that either obstruct the path or block sensor views. As a result, the system must replan locally to ensure coverage of the desired region in a safe and reliable manner. Second, changes in the surface morphology can directly affect the resolution of visual data collected at each view-pose. The preplanned route relies on viewing requirements derived from the outdated environment model. When actual surface conditions differ from the model, the inspection plan must be updated to reflect the current environment conditions.

Collectively, such morphological discrepancies translates into actions necessitating multiple mapping runs: to update the model, plan on the latest representation and finally execute the plan in the environment. As such, they contribute to an overall longer mission duration. To address these challenges, we propose a hierarchical inspection framework that aims to address the problem of routine inspection in uncertain environments. This is achieved by leveraging a priori 3D map information for the initial task planning and implementing a local reactive view planner for online replanning. Through the proposed methodology, we aim to actively address the model mismatch problem through an adaptive planning approach within a single session.

The article is structured as follows: Section~\ref{sec:related_works} presents the review on state-of-art methods and the main contributions of the article. Section~\ref{sec:prop_meth} highlights the problem statement and the proposed methodology in detail. Section~\ref{sec:setup_and_eval} and~\ref{sec:res_des} presents the evaluation setup, the experimental results and the related discussions. Finally, Sec~\ref{sec:conclusions} underscores the insights drawn from the experiments and highlights the future works of the current research.
\section{Related Works}~\label{sec:related_works}
\cite{cao2020hierarchical} presents a hierarchical inspection framework designed to provide coverage of target regions in a 3D scene. In their work, the authors propose an initial viewpoint sampling covering the desired inspection regions, followed by clustering the valid collision-free viewpoints within each environment subspace representation. Subsequently, a Travelling Salesman Problem (TSP) tour is solved and later refined to provide shortest collision-free path to cover the desired targets. Similar to this,~\cite{feng2024fc} propose a model based hierarchical approach by decomposing the 3D scene for efficient viewpoint sampling and subsequent path planning for complete coverage. Specifically, the authors leverage an input 3D point-cloud of the target scene which is then skeletonized to sample viewpoints. This is then processed to extract and execute the shortest collision-free path covering the scene. In~\cite{jin2025adaptive}, the authors present an uncertainty aware inspection framework targeted towards adaptive view planning strategy in presence of static and dynamic obstacles in the scene. Particularly, the authors leverage a priori map to compute a global view plan of the scene and then adapt it online to ensure safe navigation and occlusion-free viewing during the mission. 

While~\cite{cao2020hierarchical,feng2024fc,jin2025adaptive} demonstrate autonomous coverage of a 3D scene based on an a priori model, limited works~\cite{jin2025adaptive} address the presence of uncertainty in environments. However, these approaches often overlook a broader aspect of environmental uncertainty, namely, evolving surface morphologies. In contrast, our approach uses an a priori model to define inspection tasks and generate a global view plan. This is integrated with a reactive local view planner to maintain awareness of surface morphology and adapt in real-time, ensuring desired photogrammetric requirements are met during coverage of the inspection region.

\begin{figure*}[hbp]
    \centering
    \includegraphics[width=\linewidth]{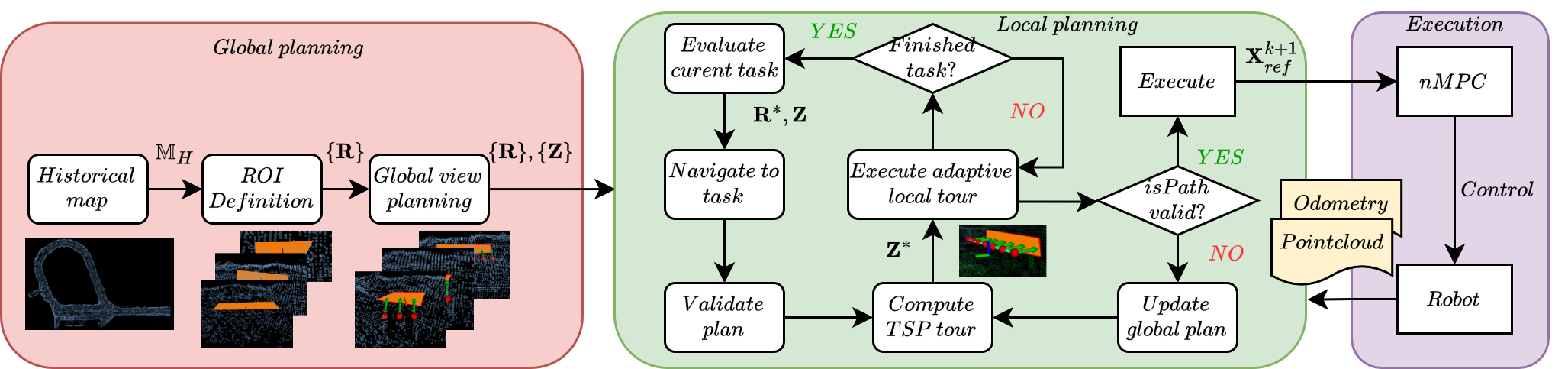}
    \caption{Framework of the proposed methodology.}
    \label{fig:framework}
\end{figure*}
\subsection{Contributions}
Building on the above discussion, we outline our key contributions that address the challenges of autonomous inspection in uncertain environments. For clarity, we define the environmental uncertainty as the incomplete or outdated information (e.g., lack of presence of obstacles, absence of knowledge on current surface morphology) available to the robot or human operator during inspection planning.

To address the above limitations, we propose a hierarchical inspection framework that combines global planning based on a historical model with local online replanning to adapt to unexpected environmental changes. Given a priori map and target inspection regions, we generate corresponding global view plans before mission start. Each inspection task is then passed to a local planner, which generates a predicted inspection route using real-time sensor data. During each task, we compute the similarity between the predicted and preplanned inspection paths, quantifying the deviation observed. When significant deviation is perceived, the system prioritizes local replanning to update the preplanned path to preserve the desired photgrammetric constraints. Throughout replanning, the system ensures awareness of the original global plan and only executes replanned segments evaluated to deviate based on the current perception of the environments. Holistically, this approach enables adaptive surface inspection in the presence of unexpected environmental changes.

Additionally, we demonstrate the efficacy of the proposed framework in real-world applications. We achieve this through field deployment experiments on a quadrupedal robot in subterranean environments. Furthermore, we evaluate the framework in large-scale missions in simulated scenarios.

\section{Methodology}\label{sec:prop_meth}
\subsection{Preliminaries}
We denote the robot's pose estimate as $\hat{\bm{X}}^k_{odom}
=[x,y,z,\phi,\theta,\psi]^T \in \mathbb{R}^6$. The camera 
horizontal and vertical Field-Of-View (FOV) are $\alpha,
\beta\in\mathbb{R}_+$, with the desired viewing distance 
$d_{view}\in\mathbb{R}_+$ and image overlaps $\gamma_H,
\gamma_V\in \mathbb{R}_+$. The input point clouds are $\bm{p}
^k=\{\bm{p}^k_i\in\mathbb{R}^3\}_{i=1}^M$. A reference view-pose 
is $\bm{X}_{ref}\in\mathbb{R}^4 |\bm{X}_{ref} = [x,y,z,\psi] $, 
and the predicted local path is $\Pi^k_{LVP}=\{\bm{X}^{k+1}_{ref},
\dots,\bm{X}^{k+N}_{ref}\}$, where $N\in\mathbb{Z}_+$ is the 
prediction horizon.
\subsection{Problem Statement}
    For a known, historical volumetric map of the environment $\mathbb{M}_H \in \mathbb{R}^3$, we denote $\bm{R}=\{\bm{v}_i\}^{M}_{i=1}|\bm{v}_i\in \mathbb{R}^3$ as a simple flat 3D polygon representing the inspection region of interest in $\mathbb{M}_H$. Given a view-pose as $\bm{x} = [x,y,z,\psi]\in\mathbb{R}^4$, we define $\bm{Z}$ as the initial set of view poses generated for $\bm{R}$ over $\mathbb{M}_H$. We denote $\mathbb{M}^{\delta} \in \mathbb{R}^3$ as the uncertainty brought about by unexpected changes in the current scene. $\mathbb{M}^{\delta}$ comprises of changes from the previously known scene, accounting for path obstructions or evolution of the mine-face topology. We denote the current 3D representation of the environment, built by the robot as it navigates the scene, as $\mathbb{M}_C \in \mathbb{R}^3$. Thus, by defining an inspection task over $\mathbb{M}_H$, the presence of $\mathbb{M}^{\delta}$ can influence the validity of $\bm{Z}$. The goal of this work is to compute and execute a an updated view plan respecting the viewing constraints in an online manner, evaluated to be necessary in presence of $\mathbb{M}^{\delta}$ and adapted to current environment conditions. Figure~\ref{fig:framework} presents the framework of the proposed methodology to address the problem of inspection in uncertain environments.

\subsection{Hierarchical Inspection Planning}

\begin{figure}[htbp]
    \centering
    \includegraphics[width=0.8\linewidth]{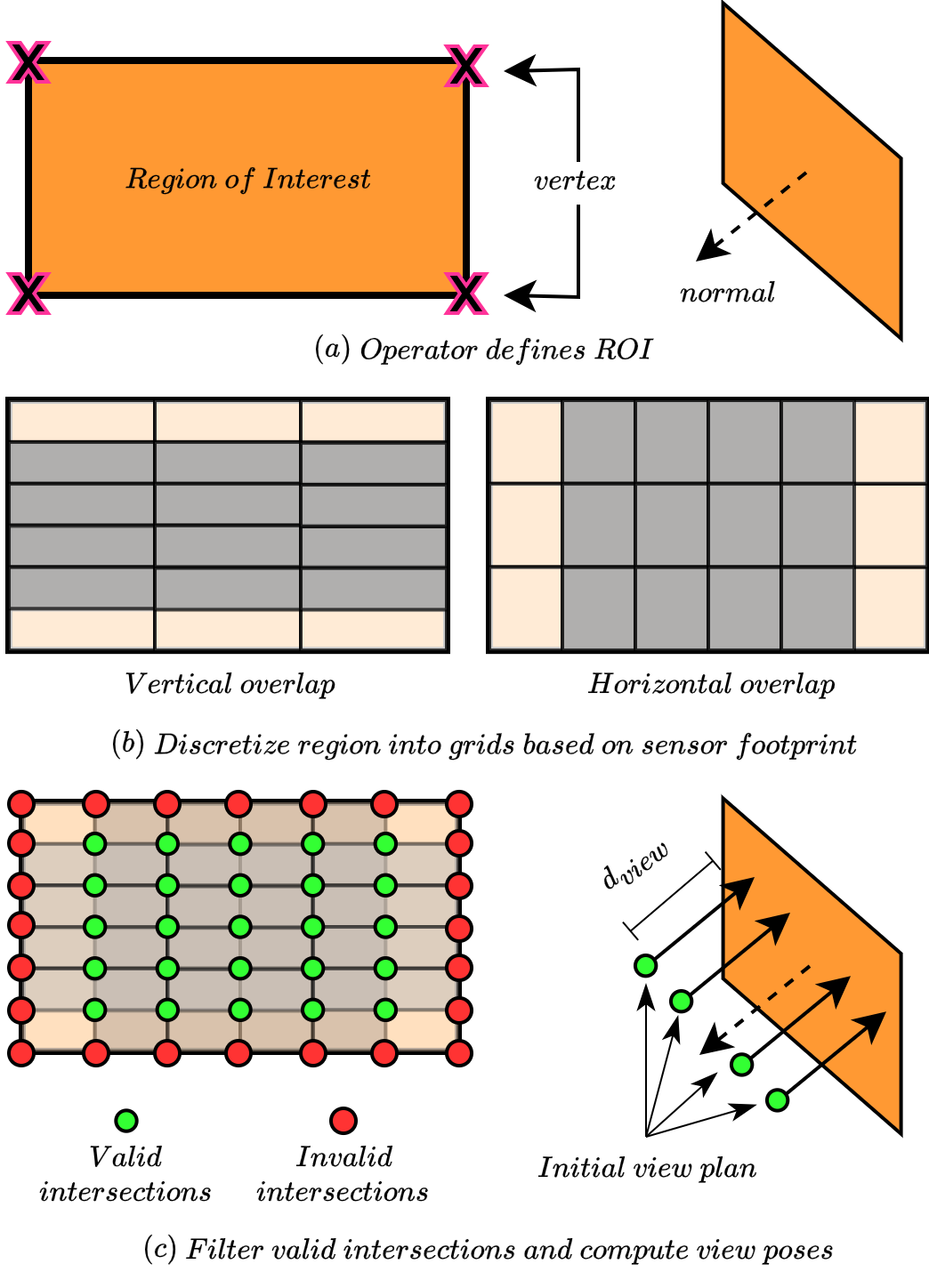}
    \caption{A visual representation of the initial view plan generation scheme for a defined region of interest.}
    \label{fig:init_vp_gen}
\end{figure}

\textbf{Global View-Planning}: Figure~\ref{fig:init_vp_gen} presents the implemented initial view plan generation scheme based on~\cite{mansouri20182d}. Subsequent to the initial ROI definition $\bm{R}$ (Fig~\ref{fig:init_vp_gen}(a)), the region is discretized into grids (Fig~\ref{fig:init_vp_gen}(b)). We consider the camera  footprint and the desired photogrammetric constraints to generate the vertical and horizontal grid lines. The distance between the grid lines is modelled based on the required overlap distance between two successive view poses. Next, we filter the grid intersections to keep only those that lie within the polygon region (Fig~\ref{fig:init_vp_gen}(c,\textit{left})). The valid intersections are then used to compute the corresponding set of view poses $\bm{Z}$. To achieve this, each valid grid intersection is projected along the direction of the normal vector of the polygon. The poses are projected to the desired viewing distance and oriented to view the ROI surface (Fig~\ref{fig:init_vp_gen}(c,\textit{right})).

\textbf{Local View Planning:} We leverage the reactive local view planner initially established in~\cite{viswanathan2024surface}. In brief, the planner operates over instantaneous 3D point cloud measurements to generate a reference view pose satisfying the desired viewing constraints. Specifically, at each planning instance, $\bm{p}^k$ is processed to evaluate the nearest 3D point on the locally observed surface $\bm{p}^k_{nn} \subseteq \bm{p}^k$. From this, the reference unit vectors capturing the ego orientation of the robot is determined. Based on the viewing constraints, $d_{view},\gamma_H,\gamma_V$, the next view-pose in sequence is computed with respect to the current position of the robot. Equation~\eqref{eqn:flip} presents the mathematical representation of the view planner. For a detailed insight on the local view planner, the readers are directed to~\cite{viswanathan2024surface}.

\begin{equation}\label{eqn:flip}
   \bm{X}^{k+1}_{ref[x,y,z]} =\hat{\bm{X}}^k_{odom[x,y,z]} + \bm{\Vec{\nu}}_x d_{insp} + \bm{\Vec{\nu}}_y d_{hov} + \bm{\Vec{\nu}}_z d_{vov}
\end{equation}
where,\begin{equation*}\label{eqn:unit_vec}
\begin{aligned}
    \Vec{\bm{\nu}}_x^k &= \frac{\bm{p}_{nn}^k - \hat{\bm{X}}^k_{odom[x,y,z]}}{||\bm{p}_{nn}^k - \bm{X}^{k}_{odom[x,y,z]}||} \\
    \Vec{\bm{\nu}}_y^k &= \Vec{\bm{\nu}}_{up}\times  \Vec{\bm{\nu}}_x^k \\
    \Vec{\bm{\nu}}_z^k &= \Vec{\bm{\nu}}_x^k \times \Vec{\bm{\nu}}_y^k \\
    d_{hov} &= 2\tan(\frac{\alpha}{2})|\bm{p}_{nn}^k - \hat{\bm{X}}^k_{odom[x,y,z]}|| \\& - 2\tan(\frac{\alpha}{2})||\bm{p}_{nn}^k - \hat{\bm{X}}^k_{odom[x,y,z]}||\gamma_H \\
    d_{vov} &= 2\tan(\frac{\beta}{2})||\bm{p}_{nn}^k - \hat{\bm{X}}^k_{odom[x,y,z]}|| \\& - 2\tan(\frac{\beta}{2})||\bm{p}_{nn}^k - \hat{\bm{X}}^k_{odom[x,y,z]}||\gamma_V
\end{aligned}
\end{equation*}
 such that, $\bm{\Vec{\nu}_x},\bm{\Vec{\nu}_y},\bm{\Vec{\nu}_z} \in \mathbb{R}^3$ are the unit view vectors of the robot with $\Vec{\bm{\nu}}_{up} = [0,0,1] $ being the unit pointing vector along the +Z-axis of the platform. Equation~\eqref{eqn:flip_yaw} presents the required yaw orientation to be maintained by the robot during local planning. 
 \begin{equation}\label{eqn:flip_yaw}
     \bm{X}_{ref[\psi]}^{k+1} =\arctan(\Vec{\bm{\nu}}_x^{k+1}(1),\Vec{\bm{\nu}}_x^{k+1}(0))
 \end{equation}

\begin{figure*}[hbtp]
    \centering
    \includegraphics[width=\linewidth]{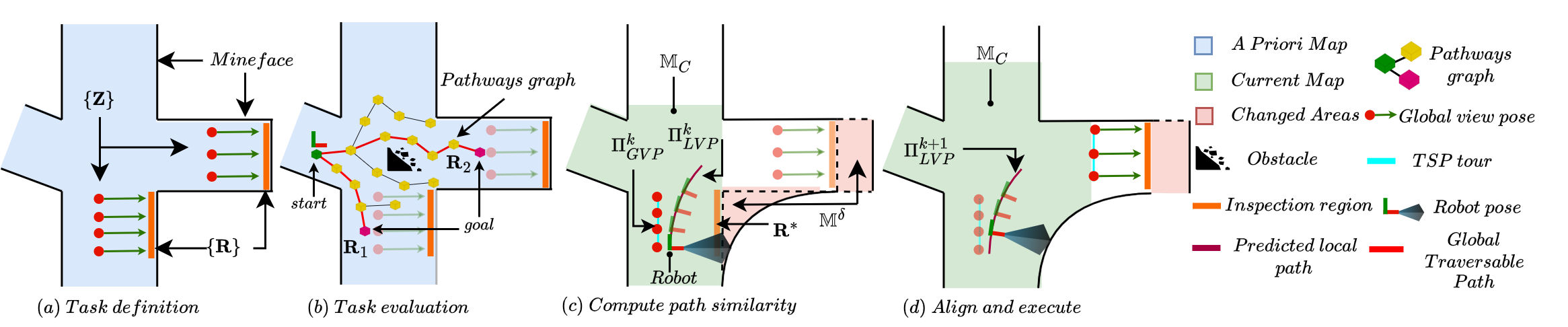}
    \caption{A visual schema of the proposed adaptive inspection planning approach in uncertain environments.}
    \label{fig:replanning}
\end{figure*}

\textbf{Planning under uncertainty:} Figure~\ref{fig:replanning} presents the pipeline for adaptive local planning in presence of environment uncertainty. Once the tasks are defined and the mission initialized, we evaluate and compute a global traversable pathway to the tasks from the current position over $\mathbb{M}_H$ (refer Fig~\ref{fig:replanning}(b)). The traversable route length is used to prioritize the closest task $\bm{R}^{*}$ respective to the robot's position. To ensure safe navigation in the scene, we segment the global path and compute and execute the new traversable route over $\mathbb{M}_C$. To generate a pathways graph, we adapt the implementation presented in~\cite{oleynikova2018safe}. If an inspection task is evaluated to be unreachable or within collision region, we update the task order and proceed to the next valid task. Though this work uses total traversable distance as the primary weighting metric to rank candidate tasks, the framework is objective-agnostic and can incorporate alternative or additional criteria to improve decision-making.

Once at target position, the effective visiting tour $\bm{Z}^{*}$, is computed from $\bm{Z}$. This is solved as a Simulated Annealing Travelling Salesman Problem (SA-TSP)~\cite{bai2023sa}. As the robot starts the tour, we extract a corresponding subset of the global inspection path $\Pi^k_{GVP} = \{\bm{x}^{*}_k\}^{k+N}_{k}|x^{*}_{k}\in \bm{Z}^{*}$. In order predict the local inspection path $\Pi^k_{LVP}$, we utilize $\Pi^k_{GVP}$ to guide the view plan generation in~\eqref{eqn:flip}. The predicted inspection path is then computed by recursively evaluating~\eqref{eqn:flip} with the updated position over the prediction horizon $N$.
We then compare this segment against the locally planned path $\Pi^k_{LVP}$ to assess the validity of $\bm{Z}$ (refer Fig~\ref{fig:replanning}(c)). Since $\Pi^k_{LVP}$ is derived from instantaneous point-cloud measurements (see~\eqref{eqn:flip}), morphological changes in the inspection surface directly influence its geometry. As a result, $\Pi^k_{LVP}$ adapts to the current environment and deviates geometrically from $\Pi^k_{GVP}$.

We model this deviation by computing the Fr\'{e}chet distance ($F_D \in \mathbb{R}_+ $) between the two paths. We integrate this measure of similarity as $\Gamma_s\in \mathbb{R}_+|\Gamma_s=[0,1]$ as defined in~\eqref{eqn:fretD}.
\begin{equation}\label{eqn:fretD}
    \Gamma_s =\frac{1}{1 + F_D(\Pi^k_{GVP},\Pi^k_{LVP})}
\end{equation}
where, $F_D$ is modelled as presented in~\cite{eiter1994computing}.

Higher values of $\Gamma_s$ indicate greater deviation between the global and local paths, implying increased environmental uncertainty compared to the prior map $\mathbb{M}_H$. To adapt planning in the presence of such uncertainty, we define a threshold $\Gamma_t \in \mathbb{R}+$, above which the local path $\Pi^k_{LVP}$ is considered as the candidate path to be tracked for the current inspection task.

Although Equation~\eqref{eqn:fretD} does not quantify uncertainty explicitly, the similarity metric $\Gamma_s$ offers an indirect measure of discrepancy in the environment, as path deviations reflect local changes in surface morphology. This introduces the challenge of keeping the inspection effort confined to the originally defined inspection region $\bm{R}^{*}$. To reconcile the updated local plan $\Pi^k_{LVP}$ with this region, we leverage an approximated global path $\hat{\Pi}_{GVP}$ to guide the replanning effort via the Kabsch method~\cite{kabsch1976solution} (refer Fig~\ref{fig:replanning}(d)). Equation~\eqref{eqn:kabsch} presents the optimal alignment used to reproject $\Pi^k_{GVP}$ in case where $\Gamma_s > \Gamma_t$.
\begin{equation}\label{eqn:kabsch}
     \hat{\Pi}_{GVP} =\begin{cases} \bm{U}^{*}\Pi^k_{GVP} + \bm{t}^{*} &,\text{if}~\Gamma_s > \Gamma_t, \\
                \Pi^k_{GVP} &,\text{otherwise} 
\end{cases}
\end{equation}
where, the optimal rotation matrix $\bm{U}^{*} \in SO(3)$ and the translation vector $\bm{t}^{*} \in \mathbb{R}^3$ are the rigid transform obtained from the Kabsch method applied to align $ \Pi^k_{GVP}$ to $ \Pi^k_{LVP}$.

Finally, the reference view plan is fed to a nonlinear model Predictive Controller (nMPC)~\cite{karlsson2022ensuring}, where the first view pose is tracked and the process~\eqref{eqn:flip}-\eqref{eqn:kabsch} repeats until the inspection task is completed. We consider the completion criteria as the robot tracking the global visitation tour $\bm{Z}^{*}$, either directly or through approximation. The pseudocode of the above process is provided in Alg~\ref{alg:gecko}.
\begin{algorithm}
    \caption{Adaptive Surface Inspection}\label{alg:gecko}

    \SetKwComment{Comment}{/* }{ */}
    \SetKwFunction{isCollisionFree}{isCollisionFree}

    \KwIn{$\mathbb{M}_{H},\alpha,\beta,\gamma_H,\gamma_V,d_{view}$}

    $\{\bm{R}\} \gets \textbf{DefineInspectionTask}(\mathbb{M}_{H})$  
    
    $\{\bm{Z}\} \gets \textbf{GenerateGlobalViewPlan}(\{\bm{R}\})$

    $\bm{R}^{*},\bm{Z} \gets \textbf{EvalInspectionTasks}(\{\bm{R}\}, \{\bm{Z}\})$
    
    $\textbf{NavigateToTask}(\bm{R}^{*})$

     \If{$\text{not}~\isCollisionFree(\bm{Z})$}{

        $\textbf{UpdatePlan}(\bm{Z})$
        
     }
    $\bm{Z}^{*} \gets \textbf{ComputeVisitingTour}(\bm{Z})$

    $\Pi^k_{GVP} \gets \textbf{ExtractGlobalSegment}(\bm{Z}^{*})$
    
    $\Pi^k_{LVP} \gets \textbf{PredictLocalPlan}(\hat{\bm{X}}^k_{odom},\bm{P}^k,\Pi^k_{GVP})$

    $\Gamma_s \gets \textbf{ComputePathSimilarity}(\Pi^k_{GVP},\Pi^k_{LVP})$

    \If{$\Gamma_s > \Gamma_t$}{
        $\hat{\Pi}^k_{GVP} \gets \textbf{ComputeAlignment}(\Pi^k_{GVP},\Pi^k_{LVP})$
        
        $\textit{nMPC} \gets \textbf{ExecuteTracking}(\Pi^k_{LVP})$
        
        $\textbf{CheckThreshold}(\hat{\Pi}^k_{GVP})$
        yout
    }\Else{
        $\textit{nMPC} \gets \textbf{ExecuteTracking}(\Pi^k_{GVP})$
        
        $\textbf{CheckThreshold}(\Pi^k_{GVP})$
    }
     $k+1 \gets k$
\end{algorithm}
\section{Setup and Evaluation}\label{sec:setup_and_eval}
\begin{figure}[htbp]
    \centering
    \includegraphics[width=\linewidth]{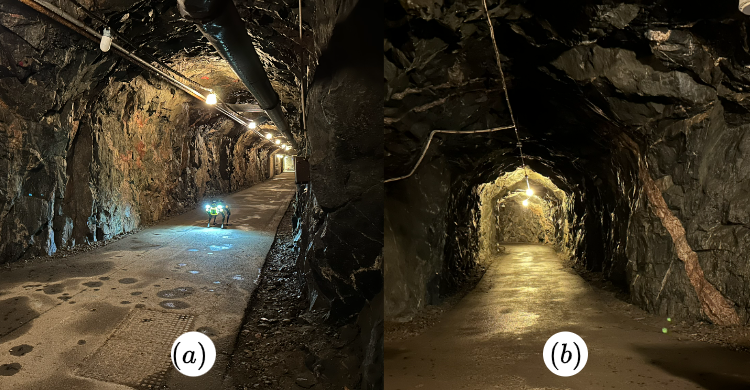}
    \caption{A visualization of the subterranean environment.}
    \label{fig:environmnt}
\end{figure}
The evaluation of the proposed methodology is carried out in real-world underground mines (refer Fig~\ref{fig:environmnt}). We evaluate along two fronts: (i) task-level inspection performance with the proposed hierarchical planner, and (ii) adaptivity to static and dynamic environmental uncertainty (e.g., surface evolution and newly introduced obstacles). To measure the quality of observation from a viewpoint, we consider the angle of incidence subtended per-pixel normal to the camera view direction. We follow the method similar to the one presented in~\cite{nakagawa2015estimating} to estimate the surface normal from the depth image. Subsequently, the viewpoint utility is computed as the mean of the cosine of the surface incidence angle, computed per valid pixel from the depth image. Throughout the runs, we consider $d_{view} = 2 \si{m}$, $\gamma_H = 0.6$, $\alpha=69.5^\circ$, $\beta=45^\circ$ and $\Gamma_t=0.5$.

The architecture is evaluated onboard Boston Dynamics (BD) Spot quadruped robot equipped with Orbbec Gemini2XL stereo-camera, Vectornav VN-100 Inertial Measurement Unit (IMU) and Ouster OS-0 3D LiDAR. We use Intel NUC embedded computational board running ROS Noetic and Ubuntu 20.04. We use Voxblox~\cite{oleynikova2017voxblox} for 3D mapping. All planning tasks are evaluated on the onboard computational unit, from loading prior map to defining and evaluating tasks and finally executing inspection of the desired inspection regions.
\section{Results and Discussions}\label{sec:res_des}
\begin{figure*}[hbp]
    \centering
    \includegraphics[width=\linewidth]{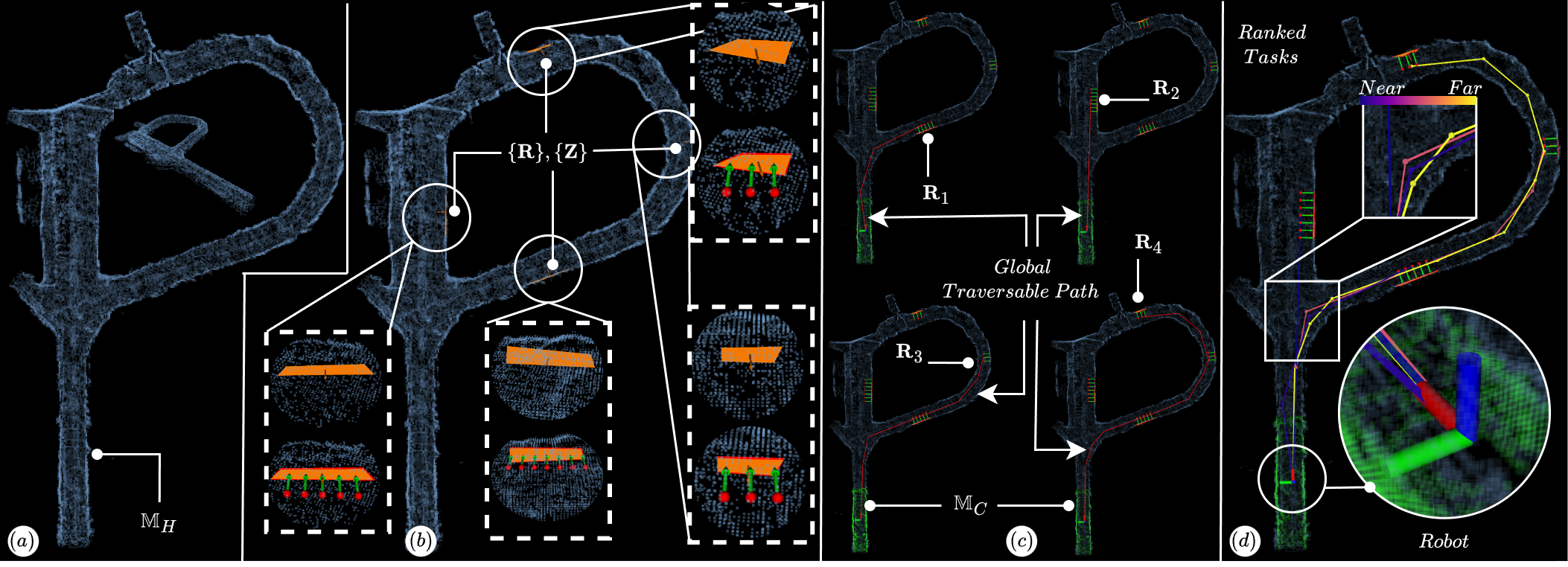}
    \caption{A collage highlighting the initial steps of the proposed hierarchical inspection framework during real-world deployment in a subterranean mine.}
    \label{fig:subt_collage}
\end{figure*}

Figure~\ref{fig:subt_collage} presents a collage of the steps taken by the hierarchical framework during deployment onboard BD SPOT in real-world subterranean mine. Fig~\ref{fig:subt_collage}(a) presents the loading of the historical map $\mathbb{M}_H$ from a previous run. The inspection tasks $\{\bm{R}\}$ are then defined within $\mathbb{M}_H$ and the global view plan $\{\bm{Z}\}$ is computed (refer Fig~\ref{fig:subt_collage}(b)). Figure~\ref{fig:subt_collage}(c) visualizes the global traversable pathway computed for each task over $\mathbb{M}_H$. Figure~\ref{fig:subt_collage}(d) visualizes these color-coded paths based on the total traversable distance from the robot's current position. Once the candidate task is evaluated, the planner proceeds to segment, recompute and execute an updated path to $\bm{R}^*$ over $\mathbb{M}_C$.

\begin{figure}[htbp]
    \centering
    \includegraphics[width=\linewidth]{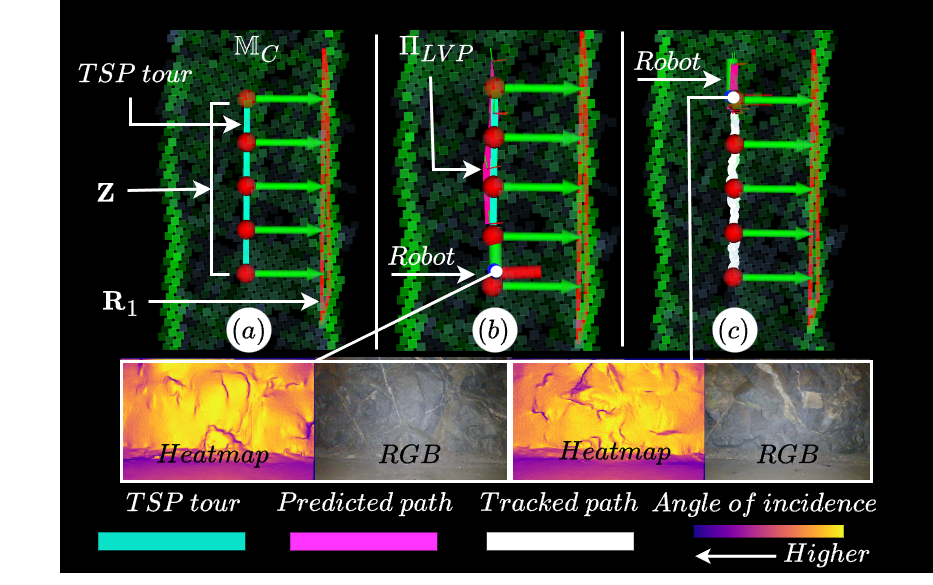}
    \caption{A visual collage capturing the inspection behaviour for region $\bm{R}_1$.}
    \label{fig:insp_r1_collage}
\end{figure}

\begin{figure}[htbp]
    \centering
    \includegraphics[width=0.9\linewidth]{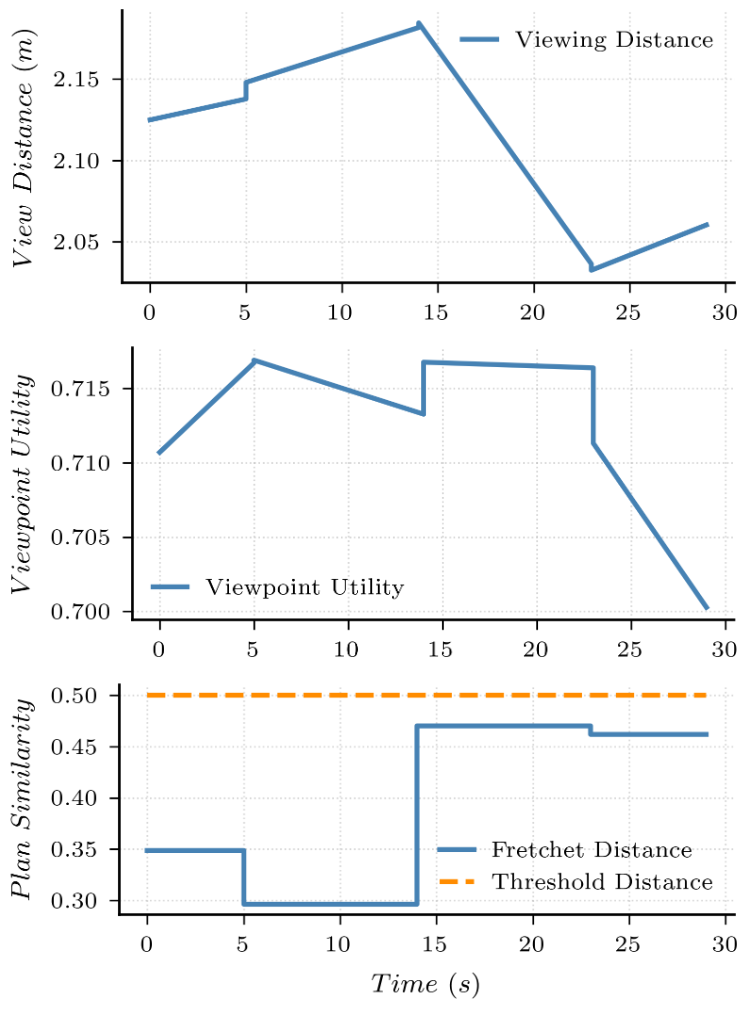}
    \caption{Inspection performance recorded for $\bm{R}_1$.}
    \label{fig:insp_r1_plots}
\end{figure}

\begin{figure}[htbp]
    \centering
    \includegraphics[width=0.9\linewidth]{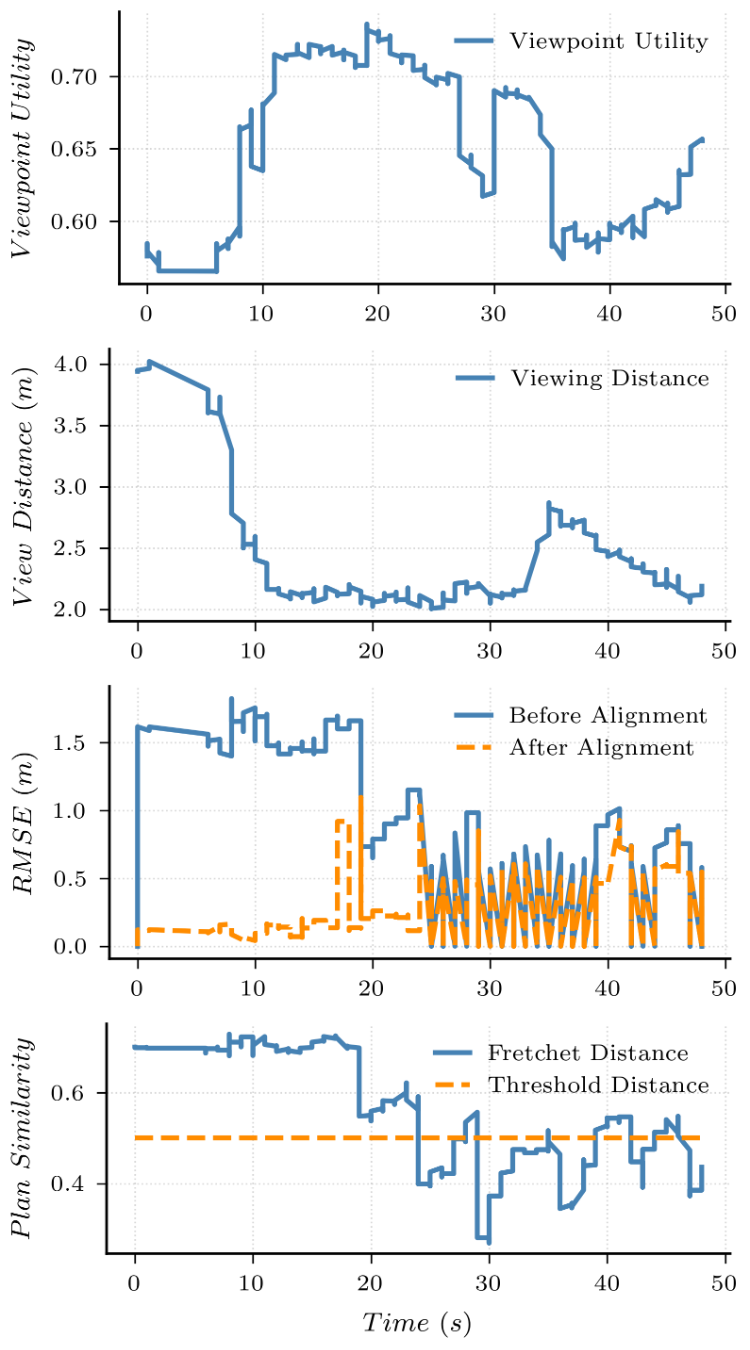}
    \caption{Inspection performance recorded for the scenario of inspection under evolving surface morphologies.}
    \label{fig:adaptive_plots}
\end{figure}

Figure~\ref{fig:insp_r1_collage} and~\ref{fig:insp_r1_plots} present a nominal scenario with inspection performance for the region $\bm{R}_1$. The task was completed in $30$ seconds with the planner evaluating $\Gamma_s$ below the defined threshold value of $0.5$ (refer Fig~\ref{fig:insp_r1_plots}(\textit{middle})). As such, the planner followed the global plan $\bm{Z}^*$. This is reflected in the profile of the predicted local inspection path (refer Fig~\ref{fig:insp_r1_collage}(b)) aligning with the global view plan during the inspection of $\bm{R}_1$. The viewpoint utility, defined earlier in Sec~\ref{sec:setup_and_eval}, records an average value of $0.71$, with values closer to $1$ indicating better orientation of the camera sensor towards the local surface.

\begin{figure*}[htbp]
    \centering
    \includegraphics[width=\linewidth]{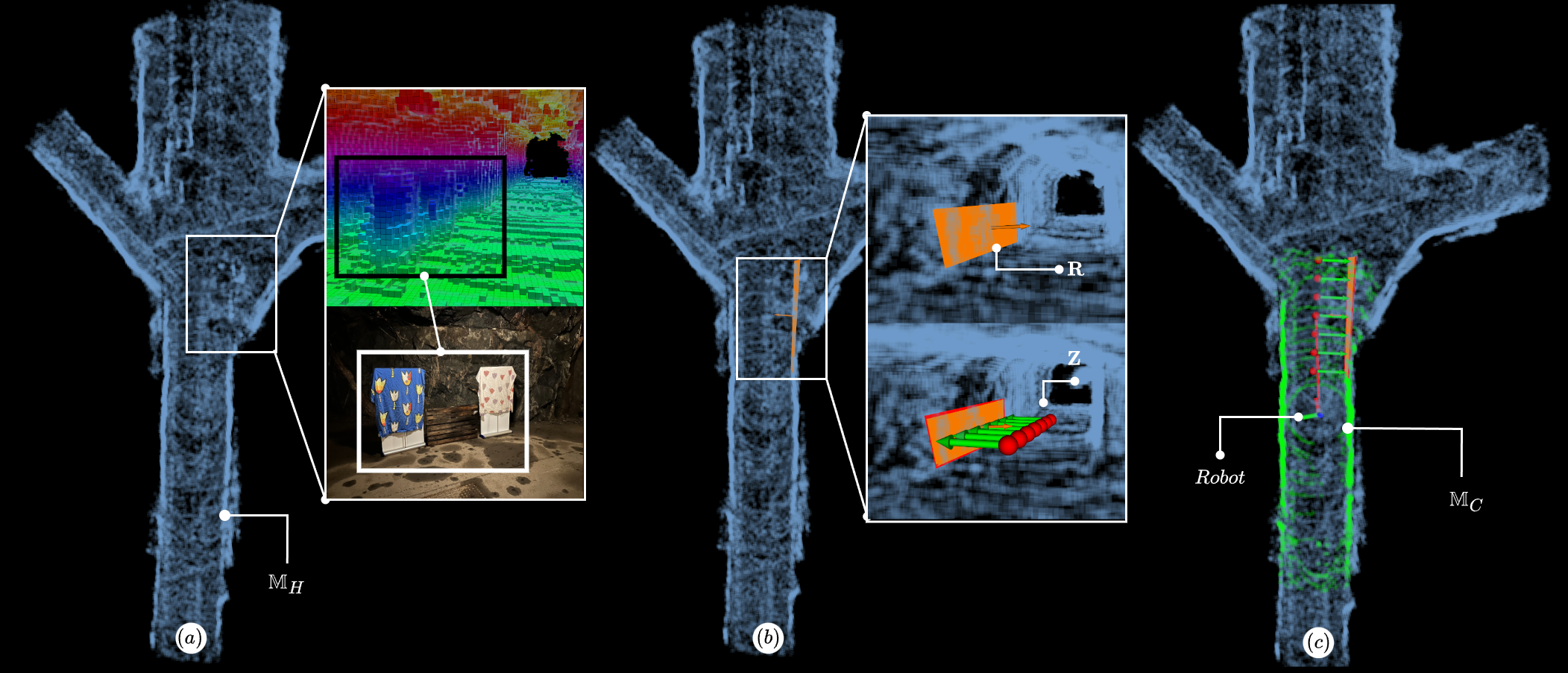}
    \caption{A visual representation of the initial inspection setup demonstrated for modified surface morphology.}
    \label{fig:evolve_morph}
\end{figure*}

For the case of addressing inspection of evolving surface morphology, Fig~\ref{fig:adaptive_plots} and~\ref{fig:evolve_morph} captures the adaptive nature of the proposed methodology. Figure~\ref{fig:evolve_morph}(a) presents the historical map built in a previous run. To simulate evolving surface morphology, we modify the scene to reflect a receding surface effect. The highlighted section in Fig~\ref{fig:evolve_morph}(a) captures the introduced objects in the scene. Once $\mathbb{M}_H$ is built, we remove the objects to simulate a receding surface. Figure~\ref{fig:evolve_morph}(b) presents the inspection region definition and subsequent computation of the global view plan. Since only a single task is defined in this scenario, the planner computes the single global pathway to the task upon mission initialization (refer Fig~\ref{fig:evolve_morph}(c)).

\begin{figure}[htbp]
    \centering
    \includegraphics[width=\linewidth]{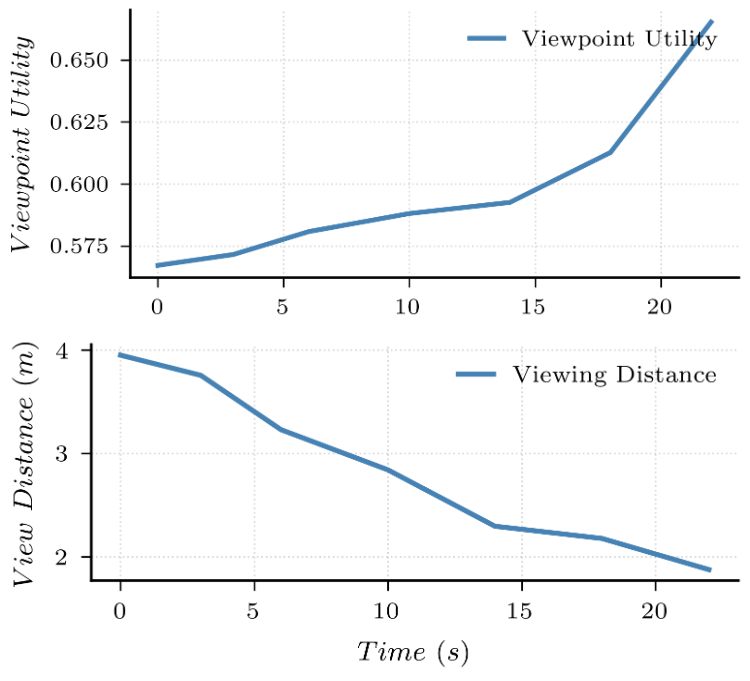}
    \caption{Inspection performance metrics recorded for a non-adaptive inspection strategy under environment uncertainty.}
    \label{fig:baseline_performance}
\end{figure}

\begin{figure*}[htb]
    \centering
    \includegraphics[width=\linewidth]{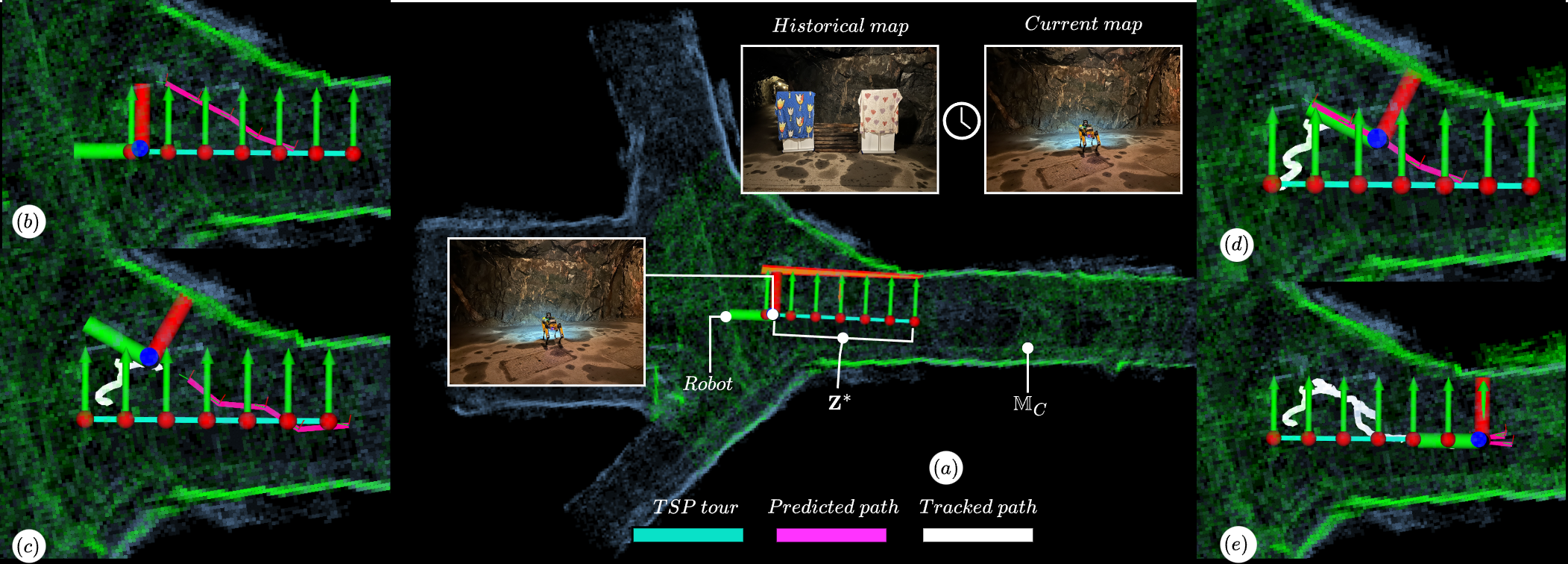}
    \caption{A collage capturing the response from the adaptive inspection strategy demonstrated during field deployment.}
    \label{fig:evolve_morph_insp}
\end{figure*}

In terms of quantitative analysis,   Figure~\ref{fig:adaptive_plots} captures the performance metrics during inspection. Particularly, the adaptive behaviour is reflected in the quick maintenance of desired viewing distance as the robot starts tracking the global view plan due to the shift in the registered surface from $\mathbb{M}_H$. This is brought into effect due to the higher $\Gamma_s$ score evaluated during inspection. From $t=20\si{s}$, the Root Mean Square Error (RMSE) between the global view plan path $\Pi_{GVP}$ and the predicted local path $\Pi_{LVP}$ before Kabsch alignment and the RMSE between $\hat{\Pi}_{GVP}$ and $\Pi_{LVP}$ after Kabsch alignment begins to converge. This indicates that the robot is moving towards inspection of a surface present in both $\mathbb{M}_C$ and $\mathbb{M}_H$. The small rise in the maintained viewing distance at ~$35\si{s}$ is due to the robot recapturing the global view plan since $\Gamma_s$ drops below the defined threshold.

Figure~\ref{fig:evolve_morph_insp} presents the adaptive inspection executed for the above scenario. In Fig~\ref{fig:evolve_morph_insp}(a), the robot's start position is highlighted alongside the main geometric difference between $\mathbb{M}_H$ and $\mathbb{M}_C$. The removed objects leave behind previously unmapped regions in $\mathbb{M}_H$. From Fig~\ref{fig:evolve_morph_insp}(a)-~\ref{fig:evolve_morph_insp}(d), the adaptive nature of the proposed methodology is shown. Once the mission starts, the planner immediately replans an updated route to meet the desired viewing constraints. This process recurses until the estimated deviation between the predicted and global view plan drops below the defined threshold. Consequently, the robot starts tracking the global tour (refer Fig~\ref{fig:evolve_morph_insp}(e)).

We compare the performance of the proposed framework against a baseline method under environmental uncertainty to highlight the improvements in view quality obtained. For baseline, we consider the robot tracking the global tour directly without the adaptive replanning. Figure~\ref{fig:baseline_performance} captures the inspection behaviour without adaptive replanning. Contrasting against the results obtained in Fig~\ref{fig:adaptive_plots}, we can directly infer that the adaptive approach maintains a higher viewpoint utility score once the replanning starts. Furthermore, the proposed methodology rapidly attempts to close the offset in the viewing distance and drops below the threshold distance in ~$15\si{s}$.

\section{Conclusions and Future Works}\label{sec:conclusions}
In this work, we presented a novel approach for routine inspection in uncertain environments. Specifically, we demonstrated an adaptive inspection planning framework capable of responding to evolving surface morphologies. We showcase our work through real-world deployments. Our method integrates a hierarchical planning structure that combines global task definition and view planning with a reactive local layer, enabling effective adaptation to dynamic surface conditions. Through the evaluated scenarios, we demonstrate the benefits of an adaptive planning approach to address environment uncertainty in a single mission. We also showcased the improvement in view quality achieved by our approach compared to a non-adaptive baseline.

For future work, we aim to extend the proposed methodology to multi-agent systems for large-scale inspection missions. Additionally, we plan to conduct more extensive field evaluations in diverse and unstructured environments to ensure robust, environment-agnostic performance.
\bibliography{bib}

@article{jin2025adaptive,
  title={Adaptive Planning Framework for UAV-Based Surface Inspection in Partially Unknown Indoor Environments},
  author={Jin, Hanyu and Xu, Zhefan and Shen, Haoyu and Han, Xinming and Ye, Kanlong and Shimada, Kenji},
  journal={arXiv preprint arXiv:2504.09294},
  year={2025}
}

@INPROCEEDINGS{viswanathan2024surface,
  author={Viswanathan, Vignesh Kottayam and Sumathy, Vidya and Kanellakis, Christoforos and Nikolakopoulos, George},
  booktitle={2024 IEEE International Conference on Robotics and Biomimetics (ROBIO)}, 
  title={A Surface Adaptive First-Look Inspection Planner for Autonomous Remote Sensing of Open-Pit Mines}, 
  year={2024},
  volume={},
  number={},
  pages={280-285},
  doi={10.1109/ROBIO64047.2024.10907551}}

@inproceedings{stathoulopoulos2023irregular,
  title={Irregular change detection in sparse bi-temporal point clouds using learned place recognition descriptors and point-to-voxel comparison},
  author={Stathoulopoulos, Nikolaos and Koval, Anton and Nikolakopoulos, George},
  booktitle={2023 IEEE/RSJ International Conference on Intelligent Robots and Systems (IROS)},
  pages={8077--8083},
  year={2023},
  organization={IEEE}
}

@article{attard2018vision,
  title={Vision-based change detection for inspection of tunnel liners},
  author={Attard, Leanne and Debono, Carl James and Valentino, Gianluca and Di Castro, Mario},
  journal={Automation in Construction},
  volume={91},
  pages={142--154},
  year={2018},
  publisher={Elsevier}
}

@inproceedings{oleynikova2017voxblox,
  title={Voxblox: Incremental 3d euclidean signed distance fields for on-board mav planning},
  author={Oleynikova, Helen and Taylor, Zachary and Fehr, Marius and Siegwart, Roland and Nieto, Juan},
  booktitle={2017 IEEE/RSJ International Conference on Intelligent Robots and Systems (IROS)},
  pages={1366--1373},
  year={2017},
  organization={IEEE}
}

@article{oleynikova2018safe,
  title={Safe local exploration for replanning in cluttered unknown environments for microaerial vehicles},
  author={Oleynikova, Helen and Taylor, Zachary and Siegwart, Roland and Nieto, Juan},
  journal={IEEE Robotics and Automation Letters},
  volume={3},
  number={3},
  pages={1474--1481},
  year={2018},
  publisher={IEEE}
}

@article{eiter1994computing,
  title={Computing discrete Fr{\'e}chet distance},
  author={Eiter, Thomas and Mannila, Heikki and others},
  year={1994},
  publisher={Technical Report CD-TR 94/64, Christian Doppler Laboratory for Expert~…}
}

@article{bai2023sa,
  title={SA-reCBS: Multi-robot task assignment with integrated reactive path generation},
  author={Bai, Yifan and Kanellakis, Christoforos and Nikolakopoulos, George},
  journal={IFAC-PapersOnLine},
  volume={56},
  number={2},
  pages={7032--7037},
  year={2023},
  publisher={Elsevier}
}

@article{nordstrom2024safety,
  title={Safety Inspections and Gas Monitoring in Hazardous Mining Areas Shortly After Blasting Using Autonomous UAVs},
  author={Nordstr{\"o}m, Samuel and Stathoulopoulos, Nikolaos and Dahlquist, Niklas and Lindqvist, Bj{\"o}rn and Tevetzidis, Ilias and Kanellakis, Christoforos and Nikolakopoulos, George},
  journal={Journal of Field Robotics},
  year={2024},
  publisher={Wiley Online Library}
}

@article{kabsch1976solution,
  title={A solution for the best rotation to relate two sets of vectors},
  author={Kabsch, Wolfgang},
  journal={Foundations of Crystallography},
  volume={32},
  number={5},
  pages={922--923},
  year={1976},
  publisher={International Union of Crystallography}
}

@article{karlsson2022ensuring,
  title={Ensuring robot-human safety for the bd spot using active visual tracking and nmpc with velocity obstacles},
  author={Karlsson, Samuel and Lindqvist, Bj{\"o}rn and Nikolakopulos, George},
  journal={IEEE Access},
  volume={10},
  pages={100224--100233},
  year={2022},
  publisher={IEEE}
}

@article{mansouri20182d,
  title={2D visual area coverage and path planning coupled with camera footprints},
  author={Mansouri, Sina Sharif and Kanellakis, Christoforos and Georgoulas, George and Kominiak, Dariusz and Gustafsson, Thomas and Nikolakopoulos, George},
  journal={Control Engineering Practice},
  volume={75},
  pages={1--16},
  year={2018},
  publisher={Elsevier}
}

@inproceedings{feng2024fc,
  title={Fc-planner: A skeleton-guided planning framework for fast aerial coverage of complex 3d scenes},
  author={Feng, Chen and Li, Haojia and Zhang, Mingjie and Chen, Xinyi and Zhou, Boyu and Shen, Shaojie},
  booktitle={2024 IEEE International Conference on Robotics and Automation (ICRA)},
  pages={8686--8692},
  year={2024},
  organization={IEEE}
}

@article{staniaszek2025autoinspect,
  title={AutoInspect: Towards Long-Term Autonomous Inspection and Monitoring},
  author={Staniaszek, Michal and Flatscher, Tobit and Rowell, Joseph and Niu, Hanlin and Liu, Wenxing and You, Yang and Gadd, Matthew and Mattamala, Mat{\'\i}as and Schutz, Alex and De Martini, Daniele and others},
  journal={IEEE Transactions on Field Robotics},
  year={2025},
  publisher={IEEE}
}

@inproceedings{cao2020hierarchical,
  title={Hierarchical coverage path planning in complex 3d environments},
  author={Cao, Chao and Zhang, Ji and Travers, Matt and Choset, Howie},
  booktitle={2020 IEEE International Conference on Robotics and Automation (ICRA)},
  pages={3206--3212},
  year={2020},
  organization={IEEE}
}

@inproceedings{nakagawa2015estimating,
  title={Estimating surface normals with depth image gradients for fast and accurate registration},
  author={Nakagawa, Yosuke and Uchiyama, Hideaki and Nagahara, Hajime and Taniguchi, Rin-Ichiro},
  booktitle={2015 International Conference on 3D Vision},
  pages={640--647},
  year={2015},
  organization={IEEE}
}
\end{document}